\def\year{2021}\relax
\documentclass[letterpaper]{article} 
\usepackage{aaai21}  
\usepackage{times}  
\usepackage{helvet} 
\usepackage{courier}  
\usepackage[hyphens]{url}  
\usepackage{graphicx} 
\usepackage{natbib}  
\usepackage{caption} 
\usepackage{amsmath}
\usepackage{mathtools}
\usepackage{amssymb}

\usepackage{caption}
\usepackage{subcaption}

\DeclareMathOperator*{\minimize}{minimize}

\title{Credit Assignment Safety Learning from Human Demonstrations}
\author{
    Ahalya Prabhakar, Aude Billard\\
}
\affiliations{
    Ecole Polytechnique Federal de Lausanne\\
    CH-1015 Lausanne, Switzerland\\
    \{ahalya.prabhakar, aude.billard\}@epfl.ch,
}

\begin{document}

\maketitle

\begin{abstract}
A critical need in assistive robotics, such as assistive wheelchairs for navigation, is a need to learn task intent and safety guarantees through user interactions in order to ensure safe task performance. For tasks where the objectives from the user are not easily defined, learning from user demonstrations has been a key step in enabling learning. However, most robot learning from demonstration (LfD) methods primarily rely on optimal demonstration in order to successfully learn a control policy, which can be challenging to acquire from novice users. 
Recent work does use suboptimal and failed demonstrations to learn about task intent; few focus on learning safety guarantees to prevent repeat failures experienced, essential for assistive robots. Furthermore, interactive human-robot learning aims to minimize effort from the human user to facilitate deployment in the real-world. As such, requiring users to label the unsafe states or keyframes from the demonstrations should not be a necessary requirement for learning.  
Here, we propose an algorithm to learn a safety value function from a set of suboptimal and failed demonstrations that is used to generate a real-time safety control filter. Importantly, we develop a credit assignment method that extracts the failure states from the failed demonstrations without requiring human labelling or prespecified knowledge of unsafe regions. Furthermore, we extend our formulation to allow for user-specific safety functions, by incorporating user-defined safety rankings from which we can generate safety level sets according to the users' preferences. By using both suboptimal and failed demonstrations and the developed credit assignment formulation, we enable learning a safety value function with minimal effort needed from the user, making it more feasible for widespread use in human-robot interactive learning tasks. 
\end{abstract}

\begin{figure}[tbh]
    \centering
    \includegraphics[width=0.5\textwidth]{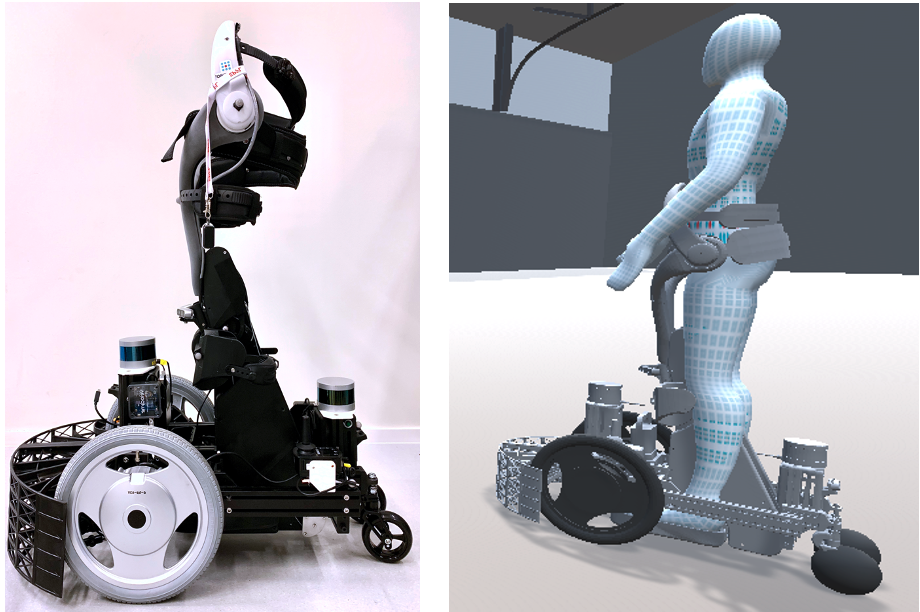}
    \caption[Qolo Robot]{A Standing Mobility Wheelchair Qolo Robot. The wheelchair's nonholonomic dynamics can be challenging to acquire high-quality demonstrations from end users for learning safety preferences and task intent. Our proposed algorithm constructs safety functions for real-time safe navigation from suboptimal and failed demonstrations.}
    \label{fig:qolo}
  \end{figure}
  
\section{Introduction}

A key challenge of developing assistive robots, such as assistive wheelchairs for nagivation, is enabling learning a user's task intent and safety limits for a task in order to ensure successful, \emph{safe} performance during task execution. For challenging tasks not easily defined by a simple objective or reward function, learning from demonstration and inverse reinforcement learning has been key for enabling robot learning even from novice users. However, LfD methods primarily rely on successful, typically optimal or expert, demonstrations in order to successfully learn a control policy or reward function, which can often be difficult to obtain, particularly when trying to extend these methods into the real-world with non-robotics experts. Furthermore, for challenging tasks, especially for tasks involving complicated robot dynamics, such as navigating a nonholonomic system (e.g., a wheelchair) or fine precision, successful demonstrations can often be difficult to obtain. This can mean that the approaches often necessarily require expert users or require users to generate a large number of attempts from which successful demonstrations can be filtered, which may be infeasible. In such cases, developing methods that enable the most information gain from the fewest number of demonstrations is critical.

Failed demonstrations can be highly informative, enabling a robot to still extract task information from them. While some recent work focusing on using suboptimal and failure demonstrations to learn task definitions (e.g., reward functions or policies), they do not extract failure states from the failure demonstrations in order to guarantee safety. This lack of guarantee is particularly important for safety-critical tasks, such as navigation and obstacle avoidance. One main challenge for doing so is the need for methods to extract the relevant task information from the provided trajectory--- in particular, determining what the cause is of the failure without requiring human knowledge or labelling of failure keyframes. While credit assignment is a challenging problem, learning safety limits without requiring user input for labeling failure states is essential for feasibly learning from humans. 

This challenge of efficient learning is further exacerbated when learning user-specific safety limits. Particularly relevant for assistive robots (e.g. wheelchair navigation), users may have different safety preferences and priorities when combining task performance and safety tolerances. 
In these scenarios, being able to effectively learn safety limits (and task intent simultaneously) efficiently is critical for deploying these methods for daily use. 
As such, the focus for developing learning algorithms in these scenarios should be focused on maximizing the information gain from each demonstration in order to minimize the effort needed from the user to learn. To do so, it requires both learning as much information as possible from a small set of demonstrations and acquiring the most informative data to aid the learning process. 

Here we present a method developed to address these main issues in the context of learning safety sets from human demonstrations that does not require predefined knowledge of the safe and unsafe regions in the workspace, for which we can provide safety guarantees for encountered failure states. As such, we focus on the following contributions here to do so: 1) developing credit assignment algorithm that utilizes human demonstrations containing optimal, suboptimal, and failure demonstrations to extract safety-guarantees without requiring human labeling of failure states; and 2) using an optional ranked demonstrations to learn user-specific safety level sets that correspond to their specific safety tolerances. This learned safety value function is used to generate a real-time resulting control safety filter that generates the optimal action with respect to task execution and safety.

\section{Related Work}

Most IRL and RL methods focus on learning a task reward function and successful task policy from a demonstration set. Few give guarantees on safety and performance \cite{argall2009survey}; though, safety has been introduced through performance bounds, with initial probabilistic bounds of the number of demonstrations needed to be within epsilon of the existing demonstrations \cite{abbeel2004apprenticeship, syed2008game}. However, they rely optimal demonstrations to enable assumptions of optimality and safety during reward and policy learning. Some recent work focusing on using suboptimal and failure demonstrations to learn what is intended to be done from them. In reinforcement learning, work using suboptimal demonstrations has been done where they initialize a policy using suboptimal demonstrations and then improve this policy using the ground truth reward signal \cite{gao2018reinforcement, taylor2011integrating}. These methods do not focus on learning what not to do from the suboptimal demonstration set, and use the learned suboptimal policy as a baseline for RL. 

Some work in IRL and Lfd have been done to extract optimal behavior from the suboptimal demonstration set. \cite{zheng2014robust, choi2019robust} both require majority of expert demonstrations in order to identify which demonstrations are suboptimal. \cite{ShiarlisIRLfailure, brown2019extrapolating} do not require majority optimal demonstrations, but show that optimal rewards can be learnt with labelled or ranked demonstrations. Few LfD work focus on learning what not to do instead. \cite{AudeDonutConf, AudeDonutJournal} require demonstrations that are clustered into two groups representing undershooting and overshooting the desired behavior. \cite{ergodiclfd2021} learns desired and undesired state spaces from a demonstration, given demonstrations labelled as success and failures. However, they do not attempt to learn safety functions or provide safety guarantees of not encountering those failure states. 

To learn a safety value function from a demonstration, we draw upon the ideas of control barrier functions. To enable safe controls and exploration, control barrier functions (CBF) \cite{ames2016control} have been used to guarantee safe actions for safety-critical systems. They use control barrier functions to generate safety control filters for task controllers \cite{ames2016control, ames2019control}, shared control \cite{zhang2020haptic}, and reinforcement learning \cite{cheng2019end}. However, these works require prespecified safety sets and corresponding CBFs in order to generate the safety-critical control. Robey et al. \cite{robey2020learning} learns the control barrier function from provided expert demonstrations. However, they still require prespecified unsafe and boundary regions in addition to the demonstrations to construct the CBF. In contrast, this work learns the CBF entirely from provided demonstrations without assuming knowledge of unsafe regions or unsafe states in the demonstration. The benefit of this formulation is that this method can be combined with any inverse reinforcement learning method (that can learn from optimal and suboptimal demonstrations) to simultaneously extract a reward function and policy from the same demonstration set. This can be combined with our learned safety controller to generate a policy that accomplishes task-relevant goals and while also guaranteeing safety. 

\section{Methods}
We describe our formulation for learning a safety value function (optionally, with user-specific safety level sets) from a set of demonstrations.   

\subsection{Problem Formulation}
We assume we have robot with state $x(t) \in \mathbb{R}^n$ and control input $u(t) \in \mathbb{R}^m$ governed by control affine dynamical system
\begin{equation}
    \dot{x}(t) = f(x(t)) + g(x(t))u(t),
\end{equation}
where $x(0) \in \mathbb{R}^n$.

Assume we have access to set of $M$ successful and unsuccessful demonstration trajectories $D = \{d_1, d_2, ... d_M\}$. Each demonstration $d_j = \{X, r\}$, consisting of a demonstration trajectory of $N$ discretized data points such that $X := \{(x_i, u_i)\}_{i=1}^N$, along with corresponding reward signals $r$ respectively that reflects demonstration quality. We require that for all failed demonstrations, reward signal is negative definite $r_f = r < 0$. The reward signal for successful demonstrations are positive semidefinite $r_s, r_b = r \geq 0$, where $r_s$ for an (optimally) safe demonstration, $r_b$ is successful but suboptimally safe (called semisafe). The reward signal values correspond to either user-specified safety ranking or fixed constants representing failure, semisafe, and optimally safe, which labels the entire demonstration. The only constraints on the reward signal are that $r_f < 0$, $r_b, r_s \geq 0$, $r_b < r_s$.

\subsection{Safety Function Optimization}
Here we show the optimization for learning the safety barrier function $h(x)$ from the set of demonstrations. To compute a valid safety value function from the demonstration trajectories, we use the following optimization problem: 
\begin{align}
    \minimize_{h \in H, \, \xi_j} \, \, \, &\| h \| + C \sum_j \xi_j
    \\ \text{subject to: } &
    \\ h(x_i) &\geq r_s, \, \forall x_i \in \bar{X}_{\text{safe}},
    \\ h(x_j) &\leq r_f + \xi_j, \, \forall x_j \in \bar{X}_{\text{unsafe}},\label{eq:unsafe-lyap}
    \\ \xi_j &> 0 , \forall j \in \mathcal{I}_+
    \\ h(x_k) &\geq r_b, \, \forall x_k \in \bar{X}_{\text{semisafe}},
    \\ q(x_k, u_k) &\geq \gamma_{\text{dyn}},  \, \forall x_i \in \bar{X}_{\text{semisafe}}  
    \\ \text{ where }&  \\q(x_k, u_k) &= \langle  \nabla h(x_k), (f(x_k) +g(x_k)u_k) + \alpha (h(x_k)) \rangle, 
\end{align}

where $\xi_j \in \mathbb{R}$ are slack variables assigned to each point in the unsafe demonstration set $\bar{X}_{\text{unsafe}}$ that relax the safety constraints. $\bar{X}_{\text{safe}}, \bar{X}_{\text{semisafe}},  \bar{X}_{\text{unsafe}} $ consists of all the trajectory points in any safe, semisafe, and failure demonstration respectively. Because all points in a failure demonstration are assigned the failure reward signal $r_f$, irrespective of if they caused the failure or not, a point may occur both in the safe demonstration set ($\bar{X}_{\text{safe}}$ or  $\bar{X}_{\text{semisafe}}$), as well as the failure set ($\bar{X}_{\text{unsafe}}$) from failure demonstrations that starts off well and end-up in a failure. As such, the slack terms are critical for enabling credit assignment, by extracting from the set of failure points, which points are safe, based on inclusion in the safe set, and which cause the failure. $C$ is a scalar penalty parameter on the slack variables and $\gamma_{dyn}$ is a hyperparameter that are set. 

We represent the barrier function $h(x)$ using RBF kernel:
\begin{align*}
  &h(x) = \sum_i^N \theta_i \phi(x,x_i) + b_i,
  \\ &\phi(x,x_i) = exp\left[\frac{-\|x-x_i\|^2}{2\sigma^2}\right]
\end{align*}

where $N$ represents the number of radial basis functions and $\phi(x,x_i)$ represents the Gaussian radial basis function, and $\theta_i$, $b_i$ are the learned parameters of the RBF.

\section{Preliminary Results}
We simulate a robotic system with state  $X(t) = [x(t), y(t)] \in \mathbb{R}^n$ and velocity-based control input $u(t) \in \mathbb{R}^m$ governed by dynamical system
\begin{align*}
  &\dot{x}(t) = u(t),
  \\ \Rightarrow &x(t) = x(t-1) + u(t)dt,
\end{align*}
where $dt$ is the time step of the system.


\subsection{Data Acquisition}
\begin{figure}[htb]
    \centering
    \includegraphics[width = 0.5\textwidth]{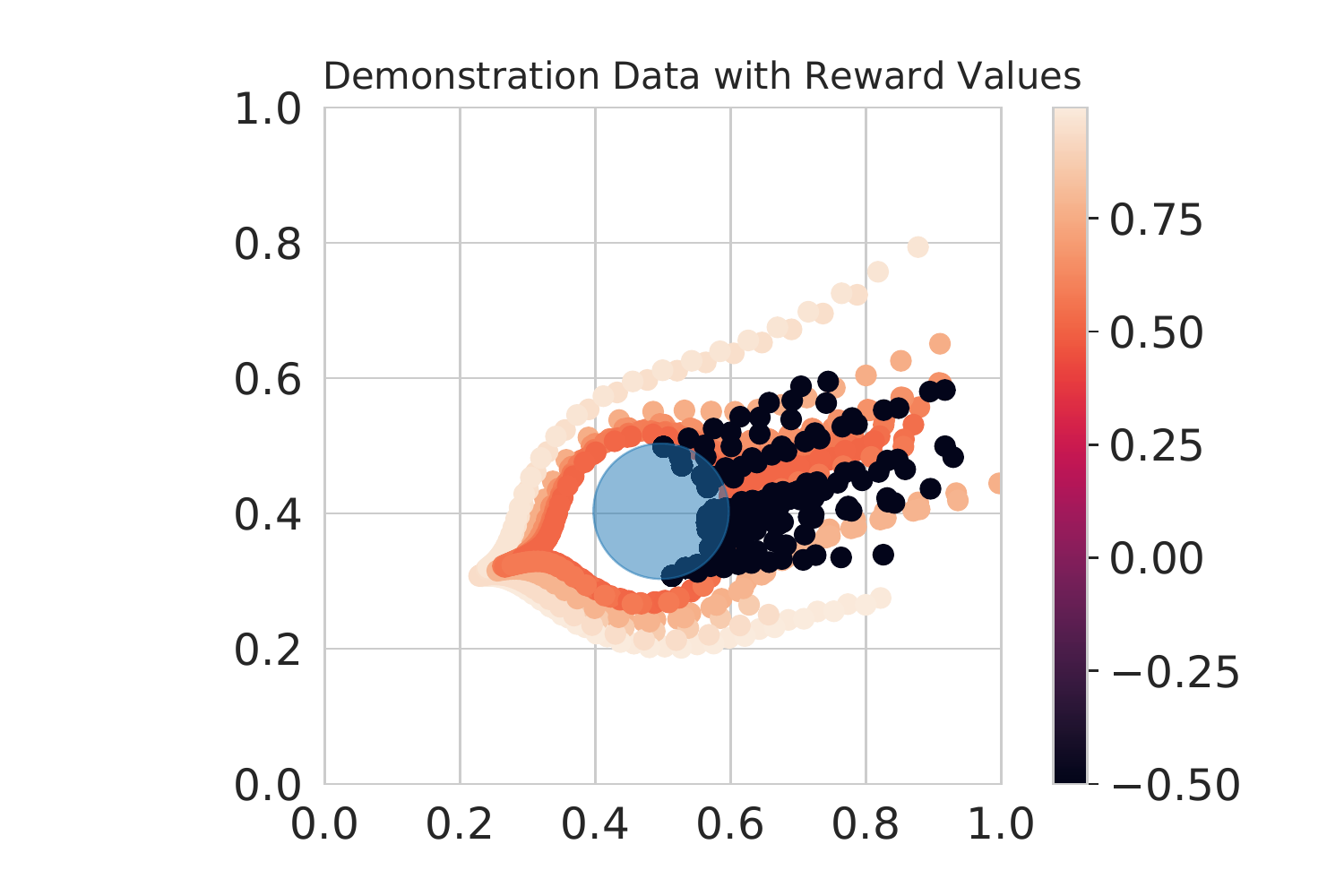}
    \caption[Demonstration Set]{Set of Demonstrations for reaching a target while avoiding an obstacle located at $(0.5,0.4)$ with radius $r_{obj}=0.1$. The set contains successful and failed demonstrations with safety reward values (reflected by the color). }
    \label{fig:obs_avoidance_data}
\end{figure}
We obtain a set of demonstrations $D = {d_1, d_2, ... d_M}$ consisting of successful (but possibly suboptimal) and failed demonstrations and an associated reward signal that represents the quality of the demonstration--- from failure demonstrations unsuccessfully attempting the task to successful demonstrations of the task itself. We define failure demonstrations as demonstrations that collide with the object and fail to reach the target as a result as shown in Figure \ref{fig:obs_avoidance_data}. We define suboptimal demonstrations here as demonstrations that are technically successful (reaching the target without colliding with the obstacle), but are too close to the obstacle for the user's preference (reflected by the safety reward value). The set of points extracted from the failure demonstrations all have the same safety reward value of $r_f = -0.5$. The set of points extracted from all successful demonstrations have safety reward values based on the the minimum distance the robot came with respect to the obstacle multiplied by the successful reward scaling $r = 5$ (i.e., $r_s = \min(distance(X_{d_i} - x_\text{obs}))*5$. All points in a single demonstration have the same safety reward value based on minimum distance it has with the obstacle. 

\subsection{Results}
\begin{figure}[tbh]
    \centering
    \includegraphics[width=0.5\textwidth]{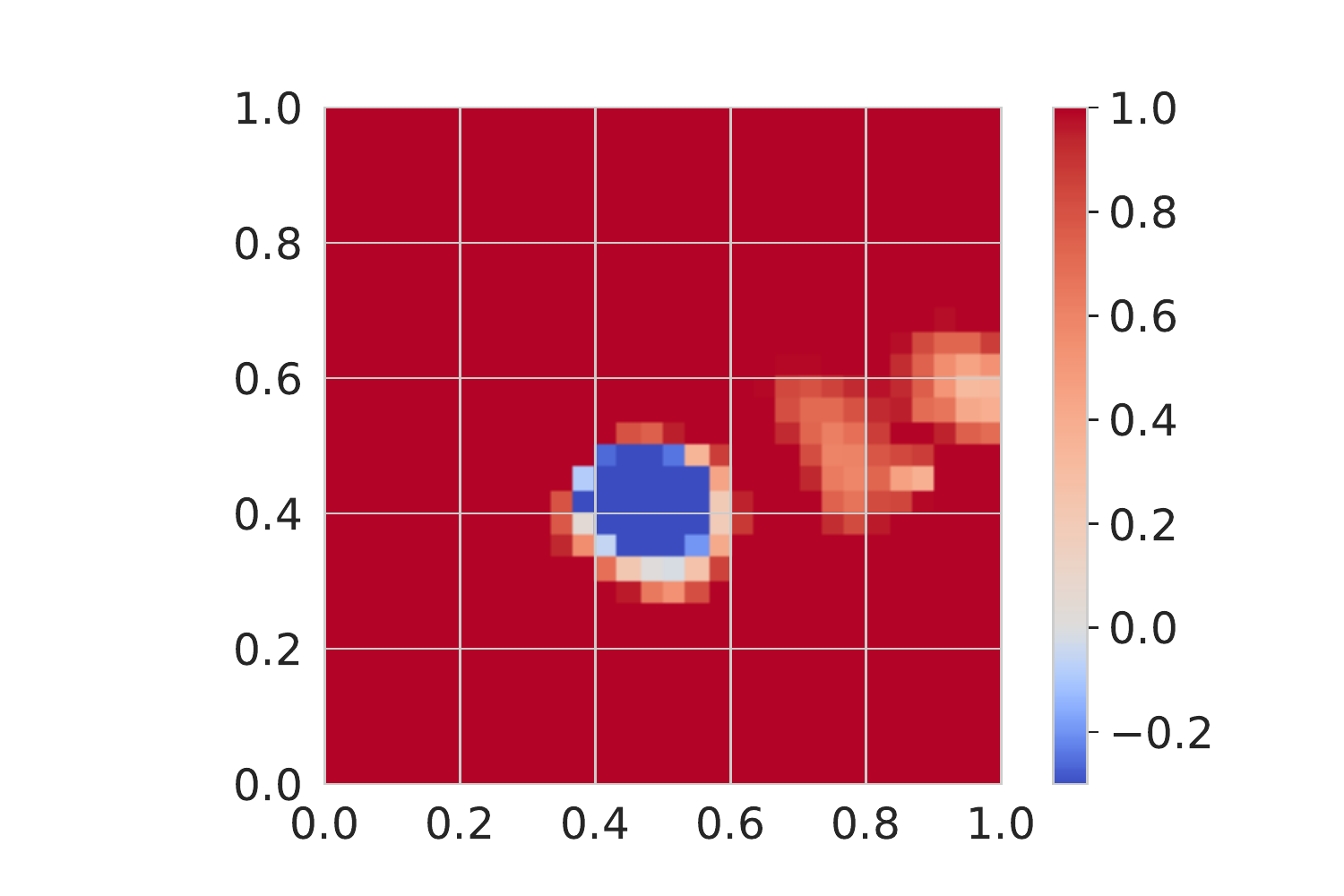}
    \caption{Learned Safety Value Function from Demonstration Set}
    \label{fig:learned_cbf}
\end{figure}

Using the proposed method, we learn a safety value function $h(x)$, reflecting the overall safety over the workspace from the demonstration set. For the learning process, we set $C = 1., \gamma_\text{dyn} = 0.1$ and $\alpha = 1$.   The learned safety function accurately extracts the unsafe region (defined as negative values in the workspace), where the obstacle is located, from the demonstration data, with prior knowledge of the location, number or shape of the object. Importantly, the failed demonstrations stop at the collision point and do not go through the obstacle itself, yet the extracted unsafe region accurately reflects the obstacle. Furthermore, all the points in the failed demonstrations have the same unsafe reward signal, but the proposed approach accurately identifies which parts of the demonstration is safe and which cause the failure. The lighter region around $(0.9,0.6$ reflects regions that are classified as less safe, as there are not enough safe demonstrations to indicate that the area is fully safe and there are some states which are only reflected in the unsafe dataset.  

\section{Planned Experiments}
  
We plan to test the full framework in a wheelchair navigation task, both in a simulated environment and and on a real wheelchair, shown in Figure \ref{fig:qolo} \cite{8594199, 9385856}. The nonholonomic dynamics and difficult controls make it challenging to generate optimal demonstrations and likelihood of failure is high during these tasks. Furthermore, users will generate demonstrations and assign a safety ranking for each based on their own preferences. Using this, we plan to develop a systematic credit assignment framework for learning and extracting safety functions from suboptimal and failed demonstrations and learn user-specific safety functions that can guarantee safety based on user preferences. These will be used to generate and test real-time safety control filters to enable real-time user control that generates the best action based on the task or user intent while still guaranteeing safety.

\section{Conclusion and Future Work}
In this work, we propose an algorithm for extracting and learning a safety value function from demonstrations with suboptimal and failed demonstrations. The algorithm enables safety credit assignment, extracting the failure states directly from the demonstration set without requiring predefined knowledge of the unsafe regions in the workspace. Furthermore, we can learn user-specific safety functions, that generate safety level sets specific to the user's particular safety tolerances. Preliminary results show successful credit assignment and safety learning, and we plan to develop formal safety guarantees for the resulting learned safety value functions. We plan to extend this work by developing an active querying algorithm that uses this framework as a initial baseline, and improves upon it by querying users for safety values for specific demonstrations and states based on the uncertainty of the learned function using information-theoretic measures as a principle for driving the active learning process.

\section{ Acknowledgments}
This work was supported by the European Research Council through the SAHR Project under the ERC Advanced Grant 741945.
\bibliography{safelearning-references.bib}
\end{document}